\documentclass{article}

\usepackage[preprint]{neurips_2024}

\usepackage[utf8]{inputenc} 
\usepackage[T1]{fontenc}    
\usepackage{hyperref}       
\usepackage{url}            
\usepackage{booktabs}       
\usepackage{amsfonts}       
\usepackage{nicefrac}       
\usepackage{microtype}      
\usepackage{xcolor}         

\usepackage{hyperref}
\usepackage{float}
\usepackage{paralist}
\usepackage{caption}

\usepackage{graphicx}

\usepackage{algorithm}
\usepackage{graphicx}
\usepackage{amsmath}
\usepackage{caption}
\usepackage{subcaption}
\usepackage{float}
\usepackage{tablefootnote}
\usepackage{multirow}
\usepackage{subcaption}
\usepackage{bbm}
\usepackage{balance}

\newcommand{\lclip}{$L_{\text{CLIP}}$ }

\newcommand{\loss}[1]{$L_{\text{#1}}$ }

\title{It's Not a Modality Gap: Characterizing and Addressing the Contrastive Gap}

\author{%
  Abrar Fahim \\
  University of Alberta\\
  \texttt{afahim2@ualberta.ca} \\
  \And
  Alex Murphy \\
  University of Alberta \\
  \texttt{amurphy3@ualberta.ca} \\
  \And
  Alona Fyshe \\
  University of Alberta \\
  \texttt{alona@ualberta.ca} \\
}

\begin{document}

\maketitle

\begin{abstract}
    Multi-modal contrastive models such as CLIP (\cite{clip}) achieve state-of-the-art performance in zero-shot classification by embedding input images and texts on a joint representational space. Recently, a  \emph{modality gap} has been reported in two-encoder contrastive models like CLIP, meaning that the image and text embeddings reside in disjoint areas of the latent space. Previous studies suggest that this gap exists due to 1) the cone effect, 2) mismatched pairs in the dataset, and 3) insufficient training. We show that, even when accounting for all these factors, and even when using the \emph{same modality}, the contrastive loss actually \emph{creates} a gap during training. As a result, We propose that the modality gap is inherent to the two-encoder contrastive loss and rename it the \emph{contrastive gap}. We present evidence that attributes this contrastive gap to low uniformity in CLIP space, resulting in embeddings that occupy only a small portion of the latent space. To close the gap, we adapt the uniformity and alignment properties of unimodal contrastive loss to the multi-modal setting and show that simply adding these terms to the CLIP loss distributes the embeddings more uniformly in the representational space, closing the gap. In our experiments, we show that the modified representational space achieves better performance than default CLIP loss in downstream tasks such as zero-shot image classification and multi-modal arithmetic. 
\end{abstract}

\maketitle

\section{Introduction}

Multi-modal models map inputs from different modalities into a unified representational space, such that semantically similar inputs from different modalities map to nearby points in the representational space. This can be practically useful because it allows transfer between modalities, but it is also more consistent with the human sensory experience, which collects information across many modalities to better understand the world. In the context of learning from paired images and text, CLIP (Contrastive Language Image Pre-training) \citep{clip} establishes a strong proof-of-concept for this reasoning. CLIP's multi-modal contrastive loss allows the model to predict the text associated with an image and vice versa. CLIP scales this approach to a very large dataset of 400M image-caption pairs, learning embeddings that cover a wide variety of visual concepts applicable to many downstream tasks.

But, while CLIP is powerful, it suffers from a \emph{modality gap} \citep{mind_the_gap}, wherein image embeddings reside in a space disjoint from that of the text embeddings. This phenomenon is also seen in multi-modal contrastive models in other domains such as medical images \citep{convirt}, videos \citep{videoclip}, amino acid sequencing (\url{https://github.com/MicPie/clasp}) and brain decoding \citep{brainscuba}. Prior work has shown that performance on downstream tasks improves when we minimize this gap, which can be achieved by simply shifting one modality's embeddings in CLIP space to change the size of the gap \citep{mind_the_gap} through projection (transforming the embeddings through some projection operation) \citep{zhou2023clippae}, or through fine-tuning \citep{geodesic}.  Recent work has additionally shown that learning latent spaces of visual embeddings via alignment with human similarity judgments improves downstream task performance \citep{MuttenthalerLDV23}. Similarly, work relating CLIP embeddings to human brain data found that representations that had reduced the gap between modalities also led to improved downstream model performance \citep{brainscuba}. Therefore, analyzing and closing this gap is a promising direction to improve upon the strong representational capacity of CLIP and its variants.

In this paper, we conducted a comprehensive study of what causes the \textit{modality gap} and simple ways to close it. We make the following contributions:

\begin{compactitem}
    \item \textbf{It's not a modality gap.} After summarizing the common purported causes of the modality gap, we perform comprehensive experiments that show that accounting for these factors does \emph{not} close the gap, suggesting that the present understanding of modality gap may be flawed.

    \item  \textbf{It's a \textit{contrastive} gap.} We present experiments that demonstrate that the gap is a byproduct of a high dimensional CLIP space, combined with contrastive loss that encourages CLIP embeddings to occupy a lower dimensional manifold relative to the latent space.

    \item \textbf{The contrastive gap can be closed.} We show that simply fine-tuning CLIP by adding a factor for uniformity and alignment can reduce the size of the gap by distributing the embeddings more uniformly throughout CLIP's latent space.

    \item \textbf{Closing the contrastive gap improves downstream performance.} Finally, we present experiments to show that closing the contrastive gap, and thereby creating more aligned and uniformly distributed representations, creates a representational space that is better for most downstream tasks, including zero-shot image classification and multi-modal embedding arithmetic. 
\end{compactitem}

\section{Background}

A multi-modal contrastive model learns a representational space in which semantically paired samples from different modalities (\emph{positive pairs}, e.g., an image and its associated caption) are closer together in the latent space compared to other randomly chosen pairs (\emph{negative pairs}, e.g., an image and the caption associated with another randomly chosen image). Contrastive learning was originally developed for the unimodal setting to learn self-supervised representations. One such popular model is SimCLR \citep{simclr}, which uses the NT-Xent (Normalized Temperature-scaled Cross entropy) loss to efficiently learn robust image representations. The NT-Xent loss is different from previous contrastive learning methods as it does not need explicit negative samples. The NT-Xent loss works by treating samples other than the positive pair in a training mini-batch as soft-negatives, pushing them away from the positives. Unlike previous contrastive methods, the NT-Xent loss also normalizes the embeddings to lie on a unit hypersphere in the representational space. 

\cite{understanding_alignment_uniformity} introduced \emph{uniformity} and \emph{alignment} factors as desirable properties of the representational space and showed that the NT-Xent loss optimizes the uniformity and alignment properties in the limit of infinite negative samples, which is equivalent to infinite batch size for NT-Xent loss.

Multiple models generalize the contrastive loss to the multi-modal setting~\citep{convirt, align, clip}. Notably, \cite{clip} scaled up vision-language contrastive learning to larger datasets to achieve impressive zero-shot classification results.

\cite{mind_the_gap} observed that a \textit{modality} gap appears in many models that use multi-modal contrastive learning. There have since been several attempts to close the gap while monitoring its effects on downstream task performance. \cite{mind_the_gap} show that altering the gap by simply translating the embeddings of one modality onto the other can positively impact zero-shot classification performance. Further, \cite{geodesic} attribute the modality gap to the absence of hard-negatives in a training mini-batch and synthetically generated hard-negative samples from existing data points, showing that training with the hard-negatives improves representational quality. In this work, we frame the gap between the multi-modal embeddings as a problem of low uniformity and show that by simply optimizing for more alignment and uniformity (properties known to be desirable in uni-modal contrastive learning), we can significantly reduce the size of the gap, while increasing the quality of the representations learned.

\subsection{Contrasting Images and Text using CLIP Loss}

CLIP is an example of a contrastive model that learns image and text embeddings. CLIP loss (\lclip) is based on NT-Xent loss. While NT-Xent loss is designed to work on data points from a single modality, \lclip is adapted to work on two different modalities of data.

In our scenario, the multi-modal dataset contains $N$ images and corresponding captions. We obtain image embeddings $E^I_j \in \mathbb{R}^d$ by passing image $I_j$ through the image encoder. Similarly, we produce the text embedding $E^T_j \in \mathbb{R}^d$ by passing caption $T_j$ through the text encoder. CLIP aims to bring image embeddings and their corresponding caption embeddings closer together in the CLIP latent space (\emph{CLIP space}) by increasing the similarity (inner product $\langle.,.\rangle$) between the corresponding embeddings. The image and text embeddings are normalized to lie on a unit hypersphere in $\mathbb{R}^d$.

The full CLIP loss is:

\begin{equation}
L_{\text{CLIP}} = -\frac{1}{2N}\sum_{j=1}^{N} \log \left[ \frac{\exp{(\langle E^I_j, E^T_j\rangle / \tau)}}{\sum_{k=1}^{N}{\exp{(\langle E^I_j, E^T_k\rangle / \tau) }}} \right] -\frac{1}{2N}\sum_{k=1}^{N} \log \left[ \frac{\exp{(\langle E^I_k, E^T_k\rangle / \tau)}}{\sum_{j=1}^{N}{\exp{(\langle E^I_j, E^T_k\rangle / \tau) }}} \right]
\label{clip_loss}
\end{equation}

where the left term \textbf{contrasts images with the texts}  ($\sum_{k=1}^{N}$ in the denominator loops over text embeddings as negatives for the $j$th image) and the right term \textbf{contrasts texts to images}  ($\sum_{j=1}^{N}$ in the denominator loops over image embeddings as negatives for the $k$th text). $\tau$ represents the temperature parameter ($\tau$ = 0.01 at the end of CLIP pre-training)

\section{The Gap in Multi-modal Contrastive Learning}

Though CLIP effectively associates images with related texts, it also creates representational spaces with a \emph{modality gap}, a phenomenon that has generated some interest. \cite{mind_the_gap} attributes this modality gap to two independent factors. First, they describe \textbf{the cone effect} of deep neural networks, in which different random initializations cause the embeddings of two encoders occupy two non-overlapping narrow cones in CLIP space. Second, they note the existence of \textbf{mismatched pairs} in the dataset, where the pairing of images and captions is incorrect for some data points. \cite{understanding_modality_gap} studied the gap in three dimensions and attributed it to conflicting uniformity and alignment terms in the contrastive loss, claiming that this leads to the existence of local minima that encourage the gap. Further, \citep{geodesic,mind_the_gap} show that the modality gap persists even after fine-tuning. In our first experiment in Section \ref{sec:gap_persists}, we will show that these factors cannot fully account for the modality gap by simulating an ideal scenario where all of the above factors are eliminated, but the gap still exists at the end of training. The goal of this work is to suggest that the \textbackslash{}emph\{modality gap\} is not caused by trying to align different modalities in a unified representational space, but instead arises as a consequence of the contrastive training that is used to align both modalities. We therefore suggest that the term \textbackslash{}emph\{contrastive gap\} better captures the underlying phenomenon.

\subsection{Measuring the gap}
To show that we have closed the gap between the embeddings of the two encoders, we must first find a way to quantify the gap. We introduce the following two metrics to measure the size and severity of the gap:

\textbf{Distance between modality centroids} (from \cite{mind_the_gap}) Given $N$ images and $N$ captions, we denote the centroid of the image embeddings as $C^I = 1/N \sum_{j=1}^{N} E^I_j$, and similarly for the centroid of the text embeddings. We compute the distance between centroids as $\lVert C^I - C^T \rVert^2$,  and note that the centroid distance can vary from $0$ to $2$. 

\textbf{Linear Separability} (from \cite{understanding_modality_gap}) is the percentage of image and text embeddings that can be distinguished by a linear classifier operating in CLIP space. We used $80\%$ of the dataset to train a linear model to classify CLIP embeddings as originating from either "image" or "text" input. We then tested the performance of the classifier on the remaining $20\%$ of the dataset and reported the accuracy. If a set of embeddings are $100\%$ linearly separable, this means that the space occupied by each modality is completely disjoint. Conversely, $50\%$ linear separability means that the image and text embeddings are overlapping in CLIP space, meaning that they occupy the same region of the latent space; i.e. there is no gap between the embeddings. To summarize, \emph{if we can effectively close the gap we will find that the distance between centroids is small and linear separability is close to 50\%}.  

\subsection{The Modality Gap Persists Even When All Factors Are Accounted For} 
\label{sec:gap_persists}
We now systematically remove the factors commonly known to contribute to the modality gap. We started with the default CLIP architecture, and used the MSCOCO \citep{mscoco} dataset. We created an idealized scenario where:

\begin{compactenum}

    \item \emph{There is only one modality.} We replaced the text encoder in CLIP with another copy of the image encoder and trained the model on pairs of images instead of text-image pairs. Thus, for this experiment the CLIP encoders are identical image encoders with different random initializations.

    \item \emph{Embeddings from the two image encoders occupy the same cone at initialization}. In our setup, after we initialize the two image encoders, we computed a fixed transformation matrix that translates the embeddings of the second image encoder to overlap with those of the first image encoder (following \cite{mind_the_gap}). Thus, the second encoder's embeddings are translated to occupy the same narrow cone as the first encoder's embeddings at initialization. This way, there is no modality gap at initialization.

    \item \emph{There are no mismatched pairs.} The positive pairs in our constructed dataset are actually identical images. This eliminated the possibility that there would be mismatched pairs in the dataset.

\end{compactenum}

We ran this experiment with a dataset of 2048 randomly selected images from the MS COCO training set. We used a batch size of 64, and the default CLIP dimensionality of 512D (i.e. Image embeddings, $E^I \in \mathbb{R}^{512}$). Since the dataset is very small, we could run the model training to completion (train loss = 0) fairly quickly (about 20k steps, or 1200 epochs).

\begin{table}[]
    \centering
    \begin{tabular}{|l|c| c|}
    \hline
                        & At initialization & After 1200 epochs  \\
                        \hline 
        Centroid distance  & 0.00        &   0.06 \\ 
        Linear separability acc. & 0.50 & 1.00 \\ 
        Contrastive loss & 4.83 & 0.00 \\
        \hline
        
    \end{tabular}
    \vspace{5mm}
    \caption{\textbf{Modality gap persists even when all factors are accounted for:} Modality gap metrics and CLIP loss values before and after training CLIP from scratch in ideal dataset conditions. \textbf{At initialization:} Distance between image and text centroids is zero and the embeddings are \emph{not} linearly separable, meaning that there is no modality gap. \textbf{After training: } Centroid distance increases slightly, but the text and image embeddings are \emph{perfectly} linearly separable. Thus, the modality gap is \emph{created} by the contrastive loss.}
    \label{tab:gap_persists}
\end{table}

We observe in Table \ref{tab:gap_persists} that even when the dataset is idealized and almost trivial to optimize on, and the model is initialized without a gap, after training to zero loss, the CLIP embeddings are perfectly linearly separable. Thus, there is a contrastive gap that is \emph{created} as a byproduct of the contrastive loss, even when the two encoders learn to align the same modality.


\subsection{Visualizing the gap in 3D CLIP}
\begin{figure*}[t!]
    \centering

     \begin{subfigure}[t]{0.22\textwidth}
        \centering
        
        \includegraphics[width=\textwidth]{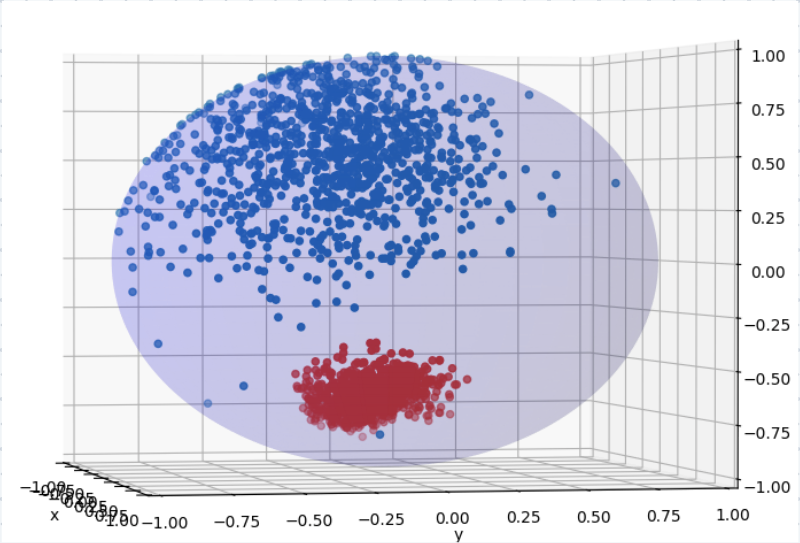}  
        \caption{\textbf{Epoch 0} \\ $I \rightarrow T$ accuracy: 0.0}
        \label{subfig:e1}
    \end{subfigure}%
    ~
    \begin{subfigure}[t]{0.22\textwidth}
        \centering
        
        \includegraphics[width=\textwidth]{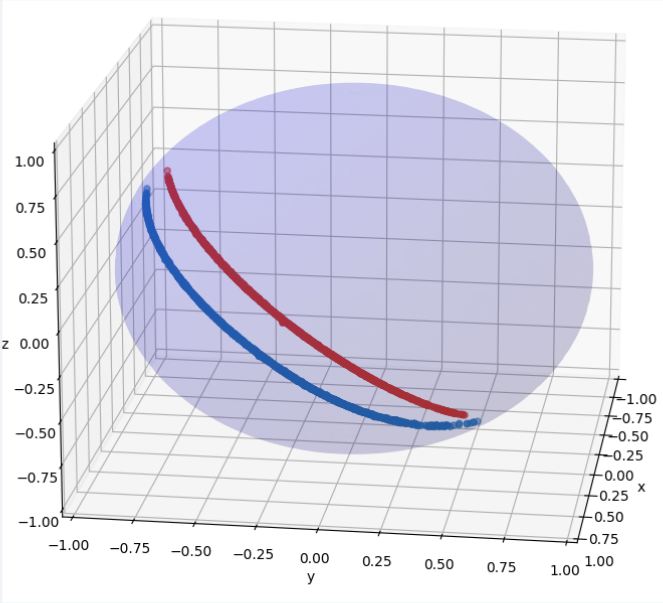} 
        \caption{\textbf{Epoch 37 }\\ $I \rightarrow T$ accuracy: 0.0}
        \label{subfig:e2}
    \end{subfigure}
    ~
    \begin{subfigure}[t]{0.22\textwidth}
        
        \centering
        \includegraphics[width=\textwidth]{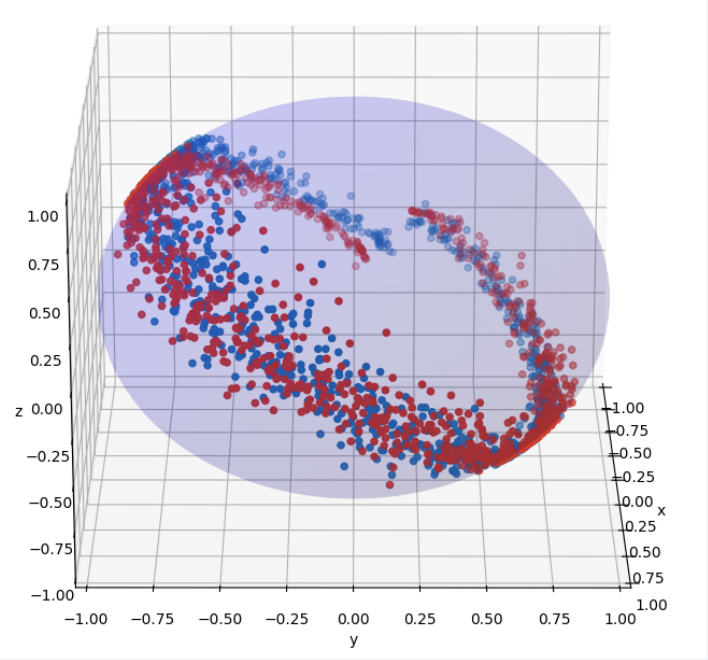} 
        \caption{\textbf{Epoch 150 }\\ $I \rightarrow T$ accuracy: 0.1}
        \label{subfig:e3}
    \end{subfigure}
    ~
    \begin{subfigure}[t]{0.22\textwidth}
        
        \centering
        \includegraphics[width=\textwidth]{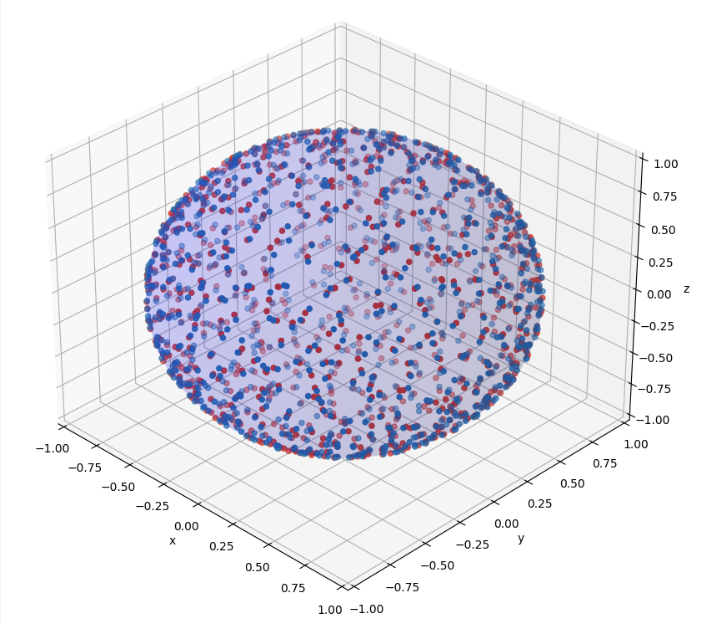} 
        \caption{\textbf{Epoch 275}\\ $I \rightarrow T$ accuracy: 0.87 }
        \label{subfig:e4}
    \end{subfigure}

    \caption{Visualizing the training stages of 3D CLIP on 1000 image-text pairs from MS COCO. Red points are image embeddings, and blue points are text embeddings. $I \rightarrow T$ accuracy represents the text retrieval accuracies: Higher $I \rightarrow T$ accuracies mean that positive pairs are well contrasted in the latent space relative to the negative pairs. The embeddings of each modality are initialized to reside in separate cones due to the cone effect, before they form \textit{arcs} and then eventually merge together as rings. Finally, they spread out to fill the sphere. }

    \label{fig:gap_closes_3d}

\end{figure*}

We showed in the previous experiment that the gap was created during training when training on 512-dimensional CLIP. However, \cite{understanding_modality_gap} shows that it is possible for CLIP loss to close the gap between embeddings by optimizing synthetically generated 3-dimensional points in Euclidean space without using a neural network.

We study the effects of CLIP loss on 3D data by optimizing a set of 1000 images and texts from MS COCO train set data using CLIP loss and the default CLIP model. In Figure \ref{fig:gap_closes_3d} we see that after training for 275 epochs, CLIP loss is able to close the contrastive gap in 3D even on real-world data. 
We report the Text-retrieval accuracies in Figure \ref{fig:gap_closes_3d} as the accuracy with which we can retrieve an input image's caption given the input image's embedding and the embeddings of all the captions in the dataset. 

We see that points on the 3D sphere are best aligned when they are evenly distributed on the sphere (as evidenced by the very high text-retrieval accuracy when Figure \ref{subfig:e4}). Therefore, we speculate that it is desirable close the contrastive gap \emph{and} to distribute the embeddings more uniformly on the unit sphere in $\mathbb{R}^d$ to improve the quality of representations. However, simply reducing CLIP dimensionality to close the contrastive gap may not be desirable, as representations retain lesser information in lower dimensions. Therefore, we propose explicitly optimizing for more uniformly distributed embeddings in high-dimensional CLIP space. We will see that when the embeddings are more uniform, the contrastive gap decreases, and in turn, the quality of the learned representations increases.


\section{Optimizing Alignment and Uniformity of CLIP Representations}

We now introduce the concepts of uniformity and alignment in the representational space. \emph{Uniformity} refers to the property of the embeddings being uniformly distributed throughout the contrastive latent space. \emph{Alignment} refers to the positive pairs being close together (aligned) in the latent space. \cite{understanding_alignment_uniformity} introduce \textit{uniformity} and \textit{alignment} as desirable properties in the unimodal contrastive representational space. i.e., image models that had higher uniformity and alignment values in their representational spaces learned better representations that led to consistent higher downstream task performance. They also showed that the unimodal contrastive loss asymptotically optimizes for uniformity and alignment with infinite negative examples. Finally, they empirically showed that, with finite negative examples, explicitly optimizing for alignment and uniformity can lead to better learned representations.

To study the effects of closing the contrastive gap, we adapt the uniformity and alignment properties from \cite{understanding_alignment_uniformity} to the multi-modal contrastive space as follows:

\begin{equation}
\text{Uniformity for Image space}: L_{\text{Uniform}}^I = \log{\left(\frac{1}{N}\sum_{j=1}^{N}\sum_{k=1}^{N} \exp{(-2 \lVert E^I_j - E^I_k \rVert^2)}\right)}
\end{equation}

We define the uniformity for text space ($L_{\text{Uniform}}^T$) similarly. Finally, the total \loss{Uniform} term is:

\begin{equation}
    L_{\text{Uniform}} = \frac{1}{2}  (L_{\text{Uniform}}^T +  L_{\text{Uniform}}^I ) \label{eq:uniform}
\end{equation}

 $L_{\text{Uniform}}^I$ and $L_{\text{Uniform}}^T$ each encourage the uniformity \emph{within} the image and text embeddings respectively. i.e., \loss{Uniform} only encourages \emph{in-modal} uniformity. The original multi-modal contrastive loss (Equation \ref{clip_loss}) does \emph{not} have any such term that constrains embeddings within each modality to be far apart. Instead, the denominators in Equation \ref{clip_loss} only push negative text samples away from positive image sample and vice versa . 

To enforce a stronger constraint on the uniformity \emph{between} negative image and text samples, we also introduce a \emph{cross-modal uniformity} term:

 \begin{equation}
    \text{Cross-modal uniformity}: L_{\text{XUniform}} = \log{\left(\frac{1}{N}\sum_{j=1}^{N}\sum_{k=1, k \neq j}^{N} \exp{(-2 \lVert E^I_j - E^T_k \rVert^2)}\right)} \label{eq:xuniform}
\end{equation}

Finally, to better align positive image-text pairs in CLIP space, we adapt the alignment term from~\cite{understanding_alignment_uniformity} to the multi-modal setting:

\begin{equation}
    L_{\text{Align}} = \frac{1}{N}\sum_{j=1}^{N} (\lVert E^I_j - E^T_j \rVert^2)
\end{equation}

Uniformity and alignment properties have been shown to be desirable properties in the uni-modal contrastive space. We validate the desirability of these two properties in the multi-modal setting. In prior works evaluating the multi-modal contrastive representational space using alignment and uniformity properties, \cite{goel_cyclip_2022} and \cite{geodesic} used cross-modality uniformity to measure uniformity in the latent space. Meanwhile \cite{al-jaff_messing_nodate} explored explicitly training multi-modal models using \loss{Uniform} (i.e., only the in-modal uniformity term) and \loss{Align}. In our work, we combine both the in-modal and cross-modal uniformity terms to encourage more uniformity in the latent space and reduce the size of the contrastive gap.

\section{Experiments}\label{sec:experiments}

We studied the effects of reducing the contrastive gap in CLIP space by optimizing for uniformity and alignment of the latent space. We fine-tuned a pre-trained CLIP model by adding the \loss{Uniform}, \loss{XUniform}, and \loss{Align} terms to the original CLIP loss (\loss{CLIP}, Equation \ref{clip_loss}). We demonstrate the effects of fine-tuning pre-trained CLIP on the following losses
\begin{compactitem}
    \item \loss{CLIP}: The default CLIP loss
    \item \loss{CLIP+Uniform+Align} (\loss{CUA}): \loss{CLIP} + \loss{Uniform} + \loss{Align}
    \item \loss{CLIP+Uniform+Align+XUniform} (\loss{CUAXU}): \loss{CLIP} + \loss{Uniform} + \loss{Align} + \loss{XUniform}
\end{compactitem}

\paragraph{Hyperparameters Overview}  For our experiments, we fine-tuned the CLIP model made available by \cite{clip}. We used the ViT-B/32 variant of the image encoder and a transformer (from OpenAI's official implementation\footnote{https://github.com/openai/CLIP}) as the text encoder. We studied the effects of fine-tuning over various sizes of CLIP space ($\mathbb{R}^d$, $d \in [32, 64, 128]$). When fine-tuning, we reduced the dimensionality of CLIP by randomly re-initializing the final linear projection layer of CLIP. For all our experiments, we fixed the temperature ($\tau$) parameter to 0.01, as $\tau$ converges to this value after CLIP pre-training.   We fine-tuned on the MS COCO \citep{mscoco} for 9 epochs. We list all hyperparameter settings in the Appendix \ref{app:hypers}

\paragraph{Experiments Structure} We first show the effects the new loss functions have on the representational space by measuring the size of contrastive gap on the MS COCO validation dataset. Then, we analyze the effects of reducing the gap by measuring image-text retreival accuracies on the MS COCO validation dataset. We then experiment to see if the same results hold for off-distribution datasets by evaluating zero-shot image classification performance on five standard datasets. Finally, we explore finer differences in the representational spaces learned by the different loss functions by reporting performance on multi-modal arithmetic.


\subsection{Effects of Reducing the Contrastive Gap on MS COCO}

First we explore the contrastive gap metrics on the MS COCO validation dataset (5k image-caption pairs) for models with different losses. Though the MS COCO dataset has 5 captions per image, we used only the first caption. We show the results for $d=128$, (other dimensionalities appear in the Appendix \ref{app:gap_metrics}). The results in Table \ref{tab:gap_uniform_vs_default} show that all the new modified CLIP losses shrink the gap. This supports our claim that increasing uniformity in CLIP space reduces the size of the contrastive gap.

\begin{figure}[t]

\begin{minipage}[b]{.49\textwidth}
    \centering

    \begin{tabular}{|l|c|c|}
       \hline
       & Linear Sep.   & Centroid    \\
       & Accuracy $(\downarrow)$ & Distance $(\downarrow)$ \\
       \hline
      \loss{CLIP}  & 1.00 &  0.66\\
       \loss{CUA} & \textbf{0.73}& \textbf{0.08}\\
       \loss{CUAXU} & 0.83& 0.14\\
       \hline
    \end{tabular}
    \captionof{table}{Gap metrics on MS COCO validation dataset. Recall: the gap closes when linear separability $\sim 0.5$ and centroid distance is small. The size of the gap is much smaller with uniformity and alignment terms included.}
    \label{tab:gap_uniform_vs_default}

\end{minipage}\hfill
\begin{minipage}[b]{.49\textwidth}
    \centering
    \begin{tabular}{|l|c|c|c|}
       \hline
       & \loss{Uniform} & \loss{XUniform} & \loss{Align}  \\
       
       \hline
       
      \loss{CLIP}  & -2.02 & -2.79 &  0.82\\
       \loss{CUA} & -3.64 & -3.68 & \textbf{0.54}\\
       \loss{CUAXU} & \textbf{-3.76} & \textbf{-3.81} & 0.69\\
       \hline
    \end{tabular}
    \captionof{table}{Final loss values for in-modality uniformity, cross-modality uniformity, and alignment on the MS COCO validation set.}
    \vspace{0.61in}
    \label{tab:mscoco_uaxu}
  \end{minipage}
\end{figure}



\begin{figure}
\begin{minipage}[b]{.49\textwidth}
    \centering
    \includegraphics[width=\textwidth]{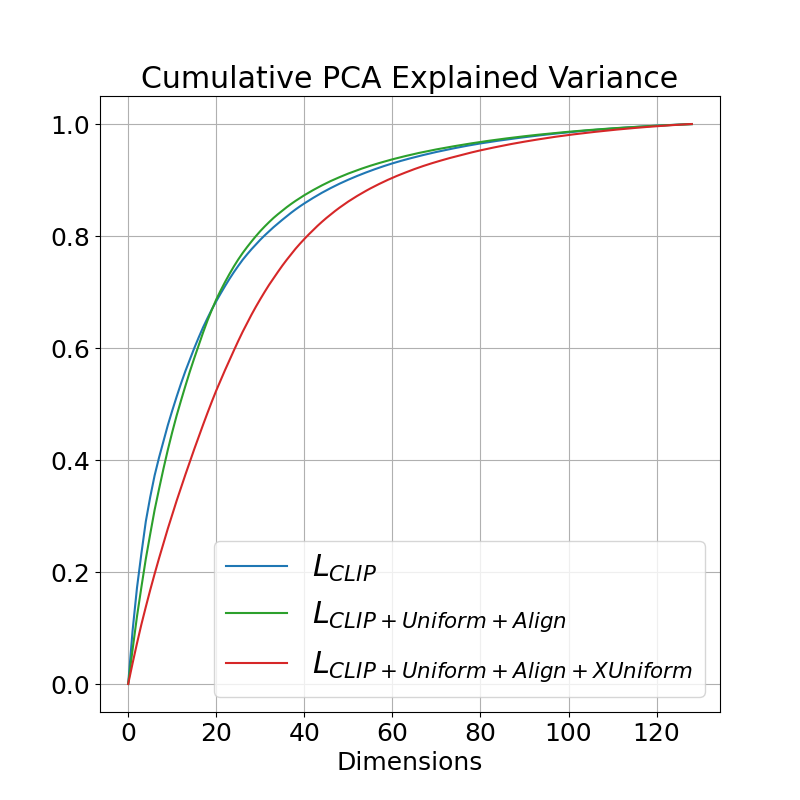}
    \caption{Explained variances for all principle components of the 128D latent space for several losses.}
    \label{fig:pca_vars}
    \vspace{0.08in}
\end{minipage}\hfill
\begin{minipage}[b]{.49\textwidth}
    \centering
    \includegraphics[width=\textwidth]{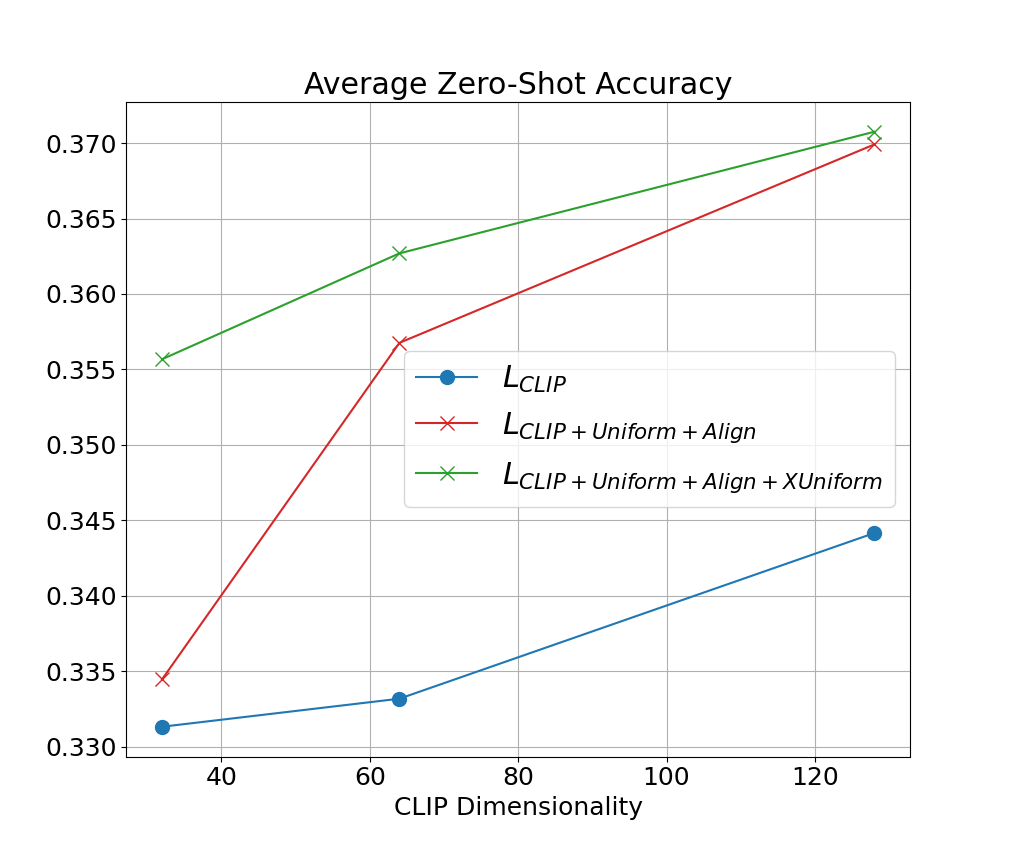}

    \caption{Average zero-shot classification accuracies for fine-tuned CLIP on the different losses. CLIP losses with uniformity and alignment terms added consistently get better zero-shot accuracies than default fine-tuned CLIP on the same dimensionality.}
    \label{fig:avg_zs_metrics}
    \label{tab:mscoco_uaxu}
  \end{minipage}
\end{figure}

Next, we look at the distribution of the image and text embeddings in CLIP space using PCA and explained variance ratios (\cite{pca}). PCA (Principle Component Analysis) is a technique used to reduce the dimensionality of a feature space by summarizing the data in a smaller set of principle components (PCs). The PCA explained variance is the amount of variance from the original data captured or "explained" by each PC. 

Figure \ref{fig:pca_vars} shows the cumulative explained variance of CLIP spaces learned using various losses. Compared to \loss{CLIP}  and \loss{CLIP+Align+Uniform}, \loss{CLIP+Align+Uniform+XUniform} has the the lowest cumulative variance across all the principle components, indicating that cross-modal uniformity encourages the embeddings to be distributed throughout the hypersphere more effectively. This is further shown in Table \ref{tab:mscoco_uaxu}, where we see that the \loss{Uniform}, \loss{XUniform}, and \loss{Align} terms are lower (i.e., embeddings are more aligned/distributed) when we add alignment and uniformity terms to the CLIP loss. 

Finally, we evaluated the quality of the representational space after reducing the contrastive gap. We used the image-text retrieval task on the MS COCO validation dataset, a standard representative task in the vision-language domain. We present the results in Table \ref{tab:mscoco_accs}.  For both the image-retrieval ($T \rightarrow I$) and the text-retrieval ($I \rightarrow T$) tasks, we see that models fine-tuned with all the loss functions perform comparably. These results suggest that it is possible to shrink the gap between the embeddings, while maintaining similar image-text retrieval performance after fine-tuning.



\begin{table}[h]
    \centering
    \begin{tabular}{|l|ccc|ccc|}

\hline
     & \multicolumn{3}{c|}{$I \rightarrow T (\uparrow)$} & \multicolumn{3}{c|}{$T \rightarrow I (\uparrow)$}\\ 
    \hline
     & Top1 & Top5  & Top10 & Top1 & Top5  & Top10 \\
       \hline \\
    
      \loss{CLIP}  & \boldmath$0.28 $ &  \boldmath$0.57$ & \boldmath$0.70$ & \boldmath$0.27$ & \boldmath$0.56$ & \boldmath$0.69$ \\
       \loss{CUA} & $0.26$  & $0.54$ & $0.68$ & $0.25$ & $0.54$ & $0.66$\\
       \loss{CUAXU} & $0.27$ & \boldmath$0.56$  & $0.69$ & \boldmath$0.27$ & \boldmath$0.56$ & \boldmath$0.69$ \\
       \hline
    \end{tabular}
    \vspace{5mm}
    
    \caption{Image to text ($I \rightarrow T$) and text to image ($T \rightarrow I$) retrieval accuracies for MS COCO. Training with uniformity/alignment maintains image-text retrieval accuracies compared to default CLIP. We report 1-standard error confidence intervals in Table \ref{tab:app_mscoco_accs} (Appendix \ref{app:image_text_retrieval}).} 
    \label{tab:mscoco_accs}
\end{table}

\subsection{Zero-Shot Transfer}

Previously, we saw that optimizing for uniformity and alignment reduces the size of the contrastive gap by encouraging the image and text embeddings to lie on higher dimensional manifolds in the unit hypersphere in $\mathbb{R}^d$. Now, we analyze the effects this has zero-shot image classification, a common downstream task associated with assessing the quality of CLIP embeddings.

We evaluate our fine-tuned CLIP models on five standard image-classification datasets: ImageNet1k (\cite{imagenet}), CIFAR 10, CIFAR 100 (\cite{cifar}), Caltech101 (\cite{caltech101}), and Describable Textures Dataset (DTD) (\cite{dtd}). We adopt the evaluation strategy of \cite{goel_cyclip_2022} and recommended by \cite{clip}: We generate prompts using the class names to form sentences like "a photo of a {class name}", "A sketch of a "class name", etc. We then pass these sentences through the text encoder to get prompt embeddings. We average all prompt embeddings to a  class embedding. We then pass the image through the image encoder and classify the image by finding the closest class via cosine similarity between the embeddings. 

Figure \ref{fig:avg_zs_metrics} shows the average zero-shot accuracies across the five datasets for each of the losses and dimensionalities. CLIP losses with alignment/uniformity consistently outperform the default CLIP loss, and XUniform adds additional benefit. In Table \ref{tab:zs_gap_metrics}, we present centroid distances between class and image embeddings, as well as uniformity loss values for image embeddings. We observe that \loss{CUA} and \loss{CUAXU} are able to learn representations with higher uniformity \emph{and} smaller contrastive gap. \loss{CUAXU} performs the best on average across all the datasets and dimensionalities in the zero-shot classification task, suggesting that lower $L_{\text{Uniform}}^I$ and lower values for centroid distance (lower contrastive gap) may lead to better representations learned.


        




       
\begin{table}[]
   \centering
   \begin{tabular}{|l|c c|}
   \hline
        &  Centroid Dist. &$L_{\text{Uniform}}^I$\\
        \hline
        \loss{CLIP}&  
        
        0.71&-1.41\\
    \loss{CUA}& \textbf{0.532}&-2.59\\
    \loss{CUAXU}& 0.59&\textbf{-2.75}\\
    \hline
    \end{tabular}
   \caption{Average contrastive gap size and image-uniformity values for 128D CLIP across all of the 5 datasets}
   \label{tab:zs_gap_metrics}
\end{table}

\subsection{Multimodal Arithmetic}

A high quality multi-modal representational space should have consistent structural representations across the learned modalities. We used SIMAT (for Semantic IMage Transformation) \citep{simat} to evaluate such relationship consistencies between CLIP image and text ebeddings. SIMAT computes a new image representation after transforming it with \emph{text delta vectors}, and retrieves the closest image to the transformed embedding. (i.e., $E^I_\text{target} = E^I_{\text{input}} + \lambda \dot (E^T_{\text{target}} - E^T_{\text{input}})$, where $\lambda$ is a hyperparameter controlling the strength of the transformation). 

Table \ref{tab:simat} shows SIMAT scores across the different loss functions. Adding uniformity and loss terms to the CLIP loss leads to increased SIMAT scores, indicating that the representational space learned with uniform / alignment terms is more consistent with arithmetic operations between modalities. This suggests that closing the contrastive gap by fine-tuning with added uniformity/alignment terms could benefit applications that rely on the geometric structure and consistent arithmetic properties in the latent space.

\begin{table}[]
    \centering
    \begin{tabular}{|l|c|}
        \hline
         & SIMAT Score  ($\uparrow$) \\
         \hline
        \loss{CLIP} & 36.02 \\
        \hline
        \loss{CLIP+Uniform+Align} & 42.18 \\
        \hline
        \loss{CLIP+Uniform+XUniform+Align} & \textbf{42.47} \\
        \hline
    \end{tabular}

    \vspace{5mm}
    \caption{SIMAT evaluation scores ($\lambda = 1$) for the different finetuning losses. Loss functions that reduce the contrastive gap produce higher SIMAT scores ($18\%$ improvement over \loss{CLIP}).}
    \label{tab:simat}
\end{table}

\section{Conclusions}\label{conclusion}

In this paper, we studied the representations learned by multi-modal contrastive learning algorithms and analyzed the contrastive gap phenomenon. We showed that eliminating all reasons commonly thought to cause the gap does \emph{not} close it. Thus, this is \textit{not a modality gap}. We instead propose the term \textit{contrastive gap} to describe this phenomenon. By studying the behaviour of the contrastive loss in 3D, we deduced that the most important factor behind the gap is low uniformity of the embeddings in the unit hypersphere. We additionally demonstrated that the contrastive gap is symptomatic of the representations lying on a lower dimensional manifold in the latent space. 

Motivated by the fact that the gap stems from low uniformity, we suggest adding explicit uniformity and alignment terms to the CLIP loss. We show that directly optimizing for uniformity and alignment in the latent space significantly reduces the gap, while forcing the image and text embeddings to lie on higher dimensional manifold on the CLIP hypersphere. This supports our claim that low uniformity in the representational space is the primary factor behind the contrastive gap.

We further show that closing the gap by simply fine-tuning CLIP with added uniformity and alignment terms improves zero-shot image classification and multi-modal arithmetic performance, while maintaining or slightly improving image-text retrieval performance.

In this work we explored the contrastive gap in the context of limited data (fine-tuning CLIP on MS COCO). In the future we would like to expand our scope and include larger datasets. Training on larger datasets could lead to more insights into the extent to which the contrastive gap closes by optimizing uniformity and alignment at scale. Finally, in our experiments, we saw that the performance in the MS COCO image-retrieval task was similar for all the losses, while the uniformity and alignment in the MS COCO space was better with the new loss factors. This suggests that there may be other measures of quality in the embedding space besides uniformity and alignment more indicative of representational quality for this task. Research into such metrics could help design more efficient loss functions while enabling better understanding of the multi-modal contrastive latent space.


\section{Broader Impacts}
\label{sec:impacts}
The work we have presented here is quite theoretical so the broader impacts are less clear. However, any model that learns from text and images has the ability to incorporate or enhance biases that exist in the training data. For example, if some captions are harmful, creating a better model to represent them may also be harmful. The images included in MS COCO also represent a biased sample of what occurs in the real world. For example, scenes from certain countries are underrepresented. This will impact any model trained on this data and could impact the utility of the model in certain deployment scenarios.x


\balance
\bibliographystyle{ACM-Reference-Format} 

\newpage
\bibliography{8.references}

\newpage

\appendix

\section{Additional Results}






\subsection{Gap Metrics on MS COCO after fine-tuning}

\label{app:gap_metrics}

We present the sizes of the contrastive gap as measured by centroid distance and linear separability accuracy on the validation dataset of MS COCO (5k image-text pairs) for each of the loss variants. Table \ref{tab:gap_uniform_vs_default} shows the size of the gap for 128-dimensional CLIP (i.e., CLIP latent space in $\mathbb{R}^d$, $d=128$).  Here, we extend those results to include gap sizes for $d \in [32, 64, 128]$. We use a fixed batch size of 64 for fine-tuning CLIP in all cases. We report 1-standard error sized confidence intervals for all values, averaged over 3 different seeds.

\begin{table}[h]
    \centering
    \begin{tabular}{|l|c|c|}
       \hline
       & Linear Separability Acc.& Centroid Distance\\
       \hline
      \loss{CLIP}  &  $0.78 \pm 0.05$ &  $0.10 \pm 0.01$\\
       \loss{CUA} & \boldmath$0.59 \pm 0.02$ & \boldmath$0.07 \pm 0.01$\\
       \loss{CUAXU} & $0.68 \pm 0.05$ & $0.12 \pm 0.03$\\
       \hline
    \end{tabular}
    \vspace{5mm}
    
    \caption{Gap metrics on MS COCO validation dataset for CLIP in $\mathbb{R}^{32}$.}
    \label{app:gaps32}
\end{table}
\begin{table}[h]
    \centering
    \begin{tabular}{|l|c|c|}
       \hline
       & Linear Separability Acc.& Centroid Distance\\
       \hline
      \loss{CLIP}  & $1.00 \pm 0.00$ & $ 0.31 \pm 0.03$\\
       \loss{CUA} & \boldmath$ 0.65 \pm 0.01$ & \boldmath$0.07 \pm 0.01$\\
       \loss{CUAXU} & $0.78 \pm 0.00$ & $0.14 \pm 0.00$\\
       \hline
    \end{tabular}
    \vspace{5mm}
    
    \caption{Gap metrics on MS COCO validation dataset for CLIP in $\mathbb{R}^{64}$.}
    \label{app:gaps64}
\end{table}
\begin{table}[h]
    \centering
    \begin{tabular}{|l|c|c|}
       \hline
       & Linear Separability Acc.& Centroid Distance\\
       \hline
      \loss{CLIP}  & $1.00 \pm 0.00$ &  $0.66 \pm 0.03$\\
       \loss{CUA} & \boldmath$0.73 \pm 0.01$ & \boldmath$0.08 \pm 0.02$\\
       \loss{CUAXU} & $0.83 \pm 0.02$ & $0.13 \pm 0.01$\\
       \hline
       
    \end{tabular}
    \vspace{5mm}
    
    \caption{Gap metrics on MS COCO validation dataset for CLIP in $\mathbb{R}^{128}$.}
    \label{app:gaps128}
\end{table}

Tables \ref{app:gaps32}, \ref{app:gaps64}, and \ref{app:gaps128} all show that the \loss{CUA} has the least gap between the embeddings (least centroid distances, and lower linear separability accuracies) across all the CLIP dimensionalities tested. The results show that increased uniformity and alignment in the representational space can reduce the size of the contrastive gap across a range of CLIP dimensionalities. Further, as the dimensionalities increase, the size of the gap also increases. For example, linear separability accuracies increase for \loss{CUA} after 9 epochs of fine-tuning as dimensionality increases from 32D to 128D. 

\begin{figure*}[h]
    \centering

     \begin{subfigure}[t]{0.3\textwidth}
        \centering
        
        \includegraphics[width=\textwidth]{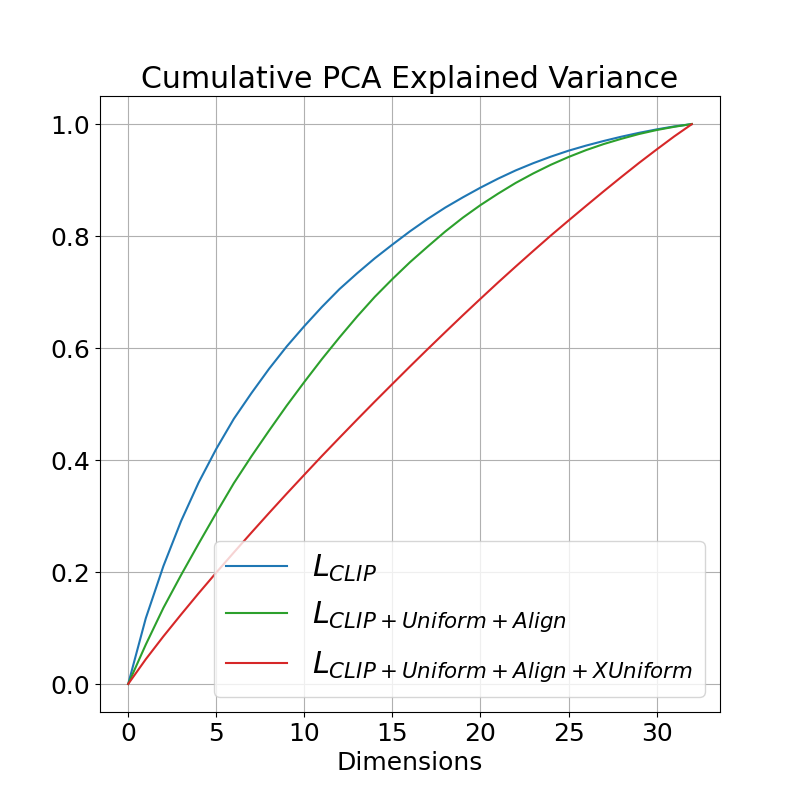}  
        \caption{\textbf{32D CLIP} }
        \label{subfig:32d_pca}
    \end{subfigure}%
    ~
     \begin{subfigure}[t]{0.3\textwidth}
        \centering
        
        \includegraphics[width=\textwidth]{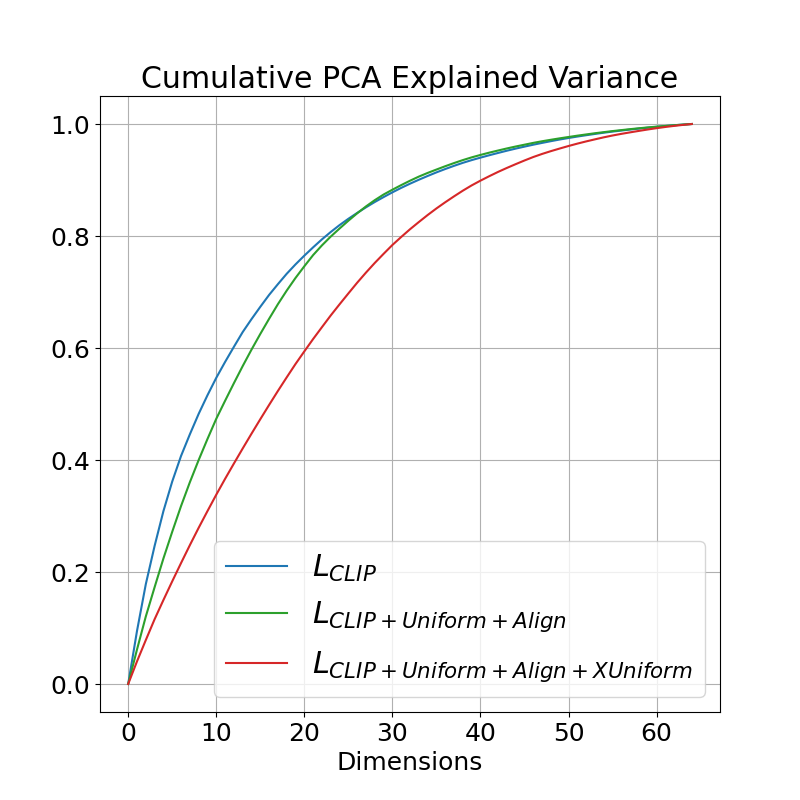}  
        \caption{\textbf{64D CLIP} }
        \label{subfig:64d_pca}
    \end{subfigure}%
    ~
     \begin{subfigure}[t]{0.3\textwidth}
        \centering
        
        \includegraphics[width=\textwidth]{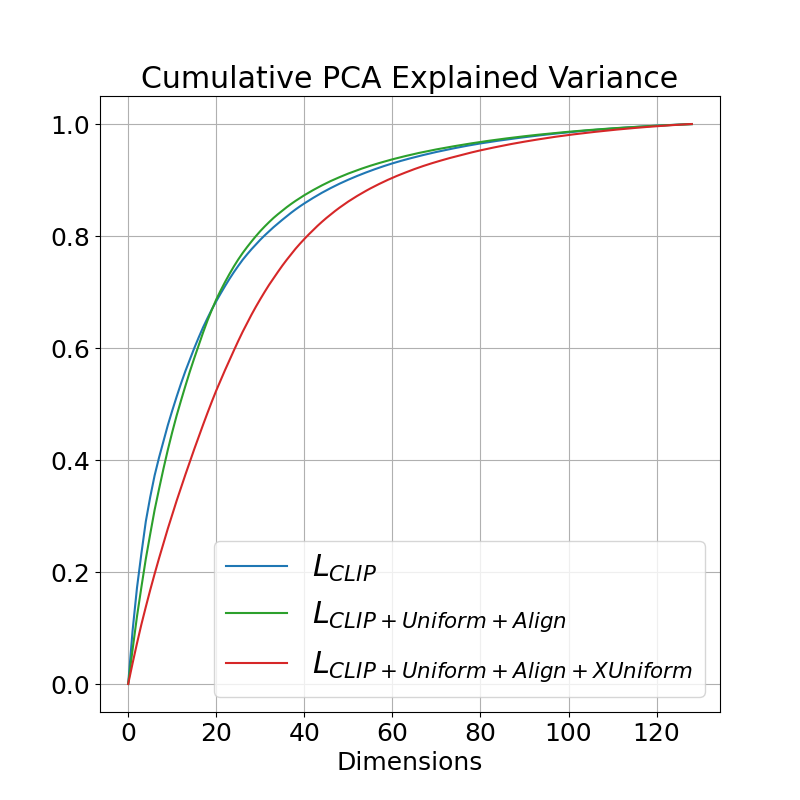}  
        \caption{\textbf{128D CLIP} }
        \label{subfig:128d_pca}
    \end{subfigure}%
    ~

    \caption{PCA explained variances for each CLIP dimensionality after fine-tuning}

    \label{fig:pca_all_dims}

\end{figure*}

Next, we look at the distribution of image and text embeddings in CLIP space using PCA and explained variance ratios. Figure \ref{fig:pca_vars} in the main text shows cumulative explained variance of CLIP spaces ( in $\mathbb{R}^{128}$) learned using the various losses. Here, we extend that figure to present the explained variance ratios for CLIP spaces in $\mathbb{R}^d, d \in [32, 64, 128]$. We show the results in Figure \ref{fig:pca_all_dims}.

From the figure, we see that \loss{CUAXU} has the lowest cumulative variance across all the principle components for all the three dimensionalities of CLIP space tested. This indicates that the the uniformity terms help to distribute the embeddings along more dimensions in the contrastive latent space in $\mathbb{R}^d$, across a wide range of $d$'s

\subsection{Image-text retrieval for MS COCO}

\label{app:image_text_retrieval}

Here, we present the results for the image-text retrieval task on MS COCO validation dataset for CLIP space in $\mathbb{R}^{128}$. We see that the image-text retrieval performance is very similar for all the three losses (within one standard error). This implies that the increased uniformity and alignment (and decreased size of contrastive gap) in the MS COCO validation space does \emph{not} correlate to better image-text retrieval performance.

\begin{table}[h]
    \centering
    \begin{tabular}{|p{0.9cm}|ccc|ccc|}

     \hline

     & \multicolumn{3}{c}{$I \rightarrow T (\uparrow)$} \vline & \multicolumn{3}{c}{$T \rightarrow I (\uparrow)$} \vline \\

      & Top1 & Top5  & Top10 & Top1 & Top5  & Top10 \\
       \hline \\
      \loss{CLIP}  & \boldmath$0.28\pm0.00 $ &  \boldmath$0.57 \pm 0.00$ & \boldmath$0.70 \pm 0.00$ & \boldmath$0.27 \pm 0.00$ & \boldmath$0.56 \pm 0.00$ & \boldmath$0.69 \pm 0.00$ \\
       \loss{CUA} & $0.26 \pm 0.00$  & $0.54 \pm 0.00$ & $0.68 \pm 0.00$ & $0.25 \pm 0.00$ & $0.54 \pm 0.00$ & $0.66 \pm 0.00$\\
       \loss{CUAXU} & $0.27 \pm 0.00$ & \boldmath$0.56 \pm 0.01$  & $0.69 \pm 0.00$ & \boldmath$0.27 \pm 0.00$ & \boldmath$0.56 \pm 0.00$ & \boldmath$0.69 \pm 0.00$ \\
       \hline
    \end{tabular}
    \vspace{5mm}
    
    \caption{Image to text ($I \rightarrow T$) and text to image ($T \rightarrow I$) retrieval accuracies for MS COCO for CLIP spaces in $\mathbb{R}^{128}$. Fine-tuning with added uniformity and alignment terms on MS COCO maintains similar levels of image-text retrieval accuracies (all values within 1 standard error of each other).}
    \label{tab:app_mscoco_accs}
\end{table}


\subsection{Zero-shot transfer}

\begin{table}[h]
    \centering
    \begin{tabular}{|c|ccccc|}
    \hline

      & CIFAR-10 & CIFAR-100 & ImageNet & DTD & Caltech101 \\ 
       
       \hline \\

       \loss{CLIP} & $0.76\pm0.02$ & $0.32\pm0.01$ & $0.13\pm0.00$ & $0.10\pm0.01$ & $0.42\pm0.01$ \\
       \loss{CUA}  &  $0.78\pm0.01$ & $0.33\pm0.00$ & $0.14\pm0.00$ & $0.10\pm0.01$ & $0.48\pm0.00$ \\
       \loss{CUAXU}  &  $0.77\pm0.02$ & \boldmath$0.35\pm0.01$ & \boldmath$0.15\pm0.00$ & $0.09\pm0.01$ & \boldmath$0.49\pm0.01$ \\
       \hline
    \end{tabular}
    \vspace{5mm}
    \caption{Zero shot classification accuracies for CLIP in $\mathbb{R}^{128}$. Error bars represent one standard error.}
    \label{app:zero_shot_accs}
\end{table}

For measuring off-distribution performance after fine-tuning on MS COCO dataset with the different CLIP loss variants, we measure the zero-shot image-classification performance on five standard image-classification datasets. We present the statistics for the validation splits of each of the datasets below (we use only the validation set for measuring zero-shot accuracies):

\begin{table}[]
    \centering
    \begin{tabular}{|c|cc|}
    \hline
       Dataset  & Test samples & Number of Classes  \\
       \hline
        CIFAR-10 & 10000 & 10\\
        CIFAR-100 & 10000 & 100\\
        ImageNet1k & 50000 & 1000\\
        DTD & 1880 & 47\\
        Caltech101 & 6084 & 102\\
        \hline
    \end{tabular}
    \vspace{5mm}
    \caption{Number of test samples, and number of classes for each of the five datasets used to measure zero-shot accuracies.}
    \label{tab:my_label}
\end{table}

We extend the results in Figure \ref{fig:avg_zs_metrics} by including individual zero-shot accuracy numbers for each of the five datasets in Table \ref{app:zero_shot_accs}.

\section{Experimental Setup}

\label{app:setup}

\subsection{Dataset}

We fine-tune on the MS COCO dataset downloaded from \citep{mscoco} \footnote{\url{https://cocodataset.org}}. We use the 2017 split, with 118k training images, and 5k validation images. Each image has 5 human-generated captions associated with it. In our experiments, we only take the first caption for each image.

\subsection{Computational Resources Used}
\label{app:compute}

 training runs, we use NVIDIA RTX A5000 GPUs, and Intel(R) Xeon(R) Silver 4210 CPUs @ 2.20GHz. One run for fine-tuning CLIP from pre-trained weights on MS COCO for 9 epochs needed about 7GB GPU memory, and needed about 3.5 hours to complete (on a single GPU). 

\subsection{Model Architecture and Hyperparameters}

\label{app:hypers}

We fine-tune the pre-trained CLIP model made available by OpenAI from Huggingface \citep{huggingface}\footnotemark. We list the model hyperparameters that we use below:

\footnotetext{{\label{hf}\url{https://huggingface.co/docs/transformers/en/model_doc/clip}}}

\begin{table}[]
    \centering
    \begin{tabular}{|c|c|}
    \hline
        Hyperparameter & Value \\
        \hline
        Image encoder & ViT/B-32 \\
        Text Encoder & Transformer (same as in \footnotemark[\value{footnote}]) \\
        Embedding dimensions & [32, 64 ,128] \\
        Temperature & 0.01 \\
        Epochs & 9 \\
        Batch size & 64 \\
        Learning rate & 1e-6 \\
        Adam beta1 & 0.9 \\
        Adam beta2 & 0.99 \\
        Adam weight decay & 0.1 \\
        Scheduler & None \\
        \hline
    \end{tabular}
    \vspace{5mm}
    \caption{Hyperparameters used for fine-tuning the CLIP models }
    \label{tab:my_label}
\end{table}

\section{Limitations}
\label{app:limitations}

In this work, we explore the properties of uniformity and alignment in the multi-modal setting, showing that added uniformity and alignment terms help to reduce the contrastive gap. One limitation of our work is that we show this at a relatively small scale by fine-tuning CLIP on the MS COCO dataset. Training CLIP from scratch on a significantly larger dataset might help gain more insights into how the contrastive loss emerges during training at a larger scale, which is important as the success of CLIP was dependent on pre-training it on a massive dataset. Another limitation is that even after optimizing for uniformity and alignment in the MS COCO space, we see that the image-text retrieval accuracies do not change significantly from that of default CLIP fine-tuning. While we argue in this work that uniformity and alignment are desirable properties in the multi-modal representational space (and directly optimize for them with our losses), the image-text retrieval accuracy numbers in the MS COCO validation set indicate that there might be other properties of the latent space that correlate to performance in this task.





    




\end{document}